\documentclass[journal]{IEEEtran}
\usepackage{amssymb}
\usepackage{amsmath}
\usepackage{graphicx}
\usepackage{caption}
\usepackage{subcaption}
\usepackage{bm} 
\usepackage{adjustbox}
\usepackage{algorithm}
\usepackage{algorithmic}
\usepackage{array}
\usepackage{cancel}
\usepackage{xfrac}
\usepackage{tikz}
\usepackage{gensymb}
\usepackage{multirow}
\usepackage{url}
\usepackage{etoolbox}
\usepackage{soul}
\usepackage[most]{tcolorbox}
\newcolumntype{C}[1]{>{\centering\arraybackslash}m{#1}}

\ifCLASSINFOpdf
\else
\fi

\begin{document}
%
\title{Rapid Gyroscope Calibration: A Deep Learning Approach}
%
%
%

\author{Yair~Stolero
        and~Itzik~Klein

\thanks{The authors are with the Hatter Department of Marine Technologies, Charney School of Marine Sciences, University of Haifa, Israel.\\ E-mails: \{ystolero@campus, kitzik@univ\}.haifa.ac.il}}

\maketitle

\begin{abstract}
Low-cost gyroscope calibration is essential for ensuring the accuracy and reliability of gyroscope measurements. Stationary calibration estimates the deterministic parts of measurement errors. To this end, a common practice is to average the gyroscope readings during a predefined period and estimate the gyroscope bias. Calibration duration plays a crucial role in performance, therefore, longer periods are preferred. However, some applications require quick startup times and calibration is therefore allowed only for a short time. In this work, we focus on reducing low-cost gyroscope calibration time using deep learning methods. We propose an end-to-end convolutional neural network for the application of gyroscope calibration. We explore the possibilities of using multiple real and virtual gyroscopes to improve the calibration performance of single gyroscopes. To train and validate our approach, we recorded a dataset consisting of 186.6 hours of gyroscope readings, using 36 gyroscopes of four different brands. We also created a virtual dataset consisting of simulated gyroscope readings. The six datasets were used to evaluate our proposed approach. One of our key achievements in this work is reducing gyroscope calibration time by up to 89\% using three low-cost gyroscopes. Our dataset is publicly available to allow reproducibility of our work and to increase research in the field.
\end{abstract}


%
\IEEEpeerreviewmaketitle

\section{Introduction}
Inertial navigation systems (INS) are commonly used in manned and autonomous platforms operating in different environments. The INS provides the navigation solution based on its inertial sensor readings \cite{Groves2013, Titterton2004}. The sensors are arranged in an inertial measurement unit (IMU) consisting of three orthogonal gyroscopes and accelerometers. INSs are popular because of their small size, low cost, high cost-effectiveness, and low power consumption. Yet, the accuracy of the INS depends heavily on the performance of its inertial sensors and their error regime, in particular in the absence of external physical sensor data or information updates\cite{farrell2008aided, engelsman2023information}. The use of low-cost IMUs further exacerbates this issue because of their large error terms and noise characteristics \cite{el2007analysis}. Recent research has examined the influence of inertial sensor errors on measurement accuracy and proposed compensation techniques to mitigate their effects {\cite{zheng2023modeling}}.

To reduce the effect of the sensor error terms on the navigation solution, stationary bias calibration of the inertial sensors is performed before the mission begins. This calibration aims to estimate the deterministic parts of the sensor errors. Once estimated, the errors are removed from the sensor readings during navigation. 
Calibration approaches can generally be divided into lab and field types. The former requires expensive equipment but it is considered to be more accurate.
Lab calibration is essential for ensuring accuracy, consistency, and reliability in sensor measurements. This process requires specialized equipment, including precision turntables and temperature chambers, to meet the stringent demands of various applications. As outlined in \cite{aggarwal2008standard}, the six-position gyroscope calibration method requires a turntable capable of precise angular velocity measurement. Another approach to IMU calibration is temperature compensation, which requires a temperature chamber. Sensor readings can drift because of temperature variations, therefore a temperature sensor is used to record these changes, allowing the creation of a compensation model that adjusts the sensor readings accordingly \cite{cui2019mems}. A hybrid calibration technique for fiber optic gyroscopes using low-cost IMUs is presented in \cite{xu2019novel}. In \cite{choi2010calibration}, gyroscope calibration is carried out using a pendulum. Additional in-lab calibration methods include self-calibration through factorization \cite{hwangbo2013imu}, gyroscope denoising \cite{engelsman2023data}, IMU hand calibration {\cite{al2023imu}}, self-calibration using machine learning {\cite{tritschler2024meta}}, and deep learning approaches \cite{chen2022towards}.

Yet, the equipment needed to perform high-accuracy calibration is not always available. To address this problem, field calibration methods have been developed, including some tailored for low-cost IMUs. For example, Lasmadi showed how to mitigate sensor errors and bias compensation using a Kalman filter and zero velocity compensation \cite{lasmadi2017inertial}. The 3-axis rotations calibration is performed by rotating the gyroscope around each of its three axes and recording the output to calibrate the bias, scale factor, and misalignment errors \cite{wang2021efficient}. In \cite{bhatia2016development}, an analytical method is developed. The IMU is moved only by hand to various locations, then using certain analytical equations, the IMU is calibrated. The above are some ideas for calibrating gyroscopes, but when considering only the bias as the main error term, the fastest reliable method is the zero-order calibration \cite{kirkko2012bias}, where regardless of the gyroscope orientation, a predefined sequence of readings is averaged to estimate the bias. A more comprehensive view of calibration methods can be found in \cite{poddar2017comprehensive, harindranath2023systematic}.

In parallel with the advances in field calibration approaches for a single IMU with model-based methods, important breakthroughs occurred in the field of multiple inertial measurement units (MIMU) and inertial sensing based on artificial intelligence. In a MIMU setup, multiple IMUs are rigidly connected and aligned with one another \cite{larey2020multiple}. A broad overview of the topic is presented in \cite{nilsson2016inertial}. Applying a data fusion algorithm to the output of MIMU, two objectives can be accomplished: (a) the ability to detect outlier measurements, and (b) a general reduction in errors, specifically the IMU noise \cite{libero2024augmented}. Skog \emph{et al}. demonstrated how these goals are achieved in a massive MIMU structure \cite{skog2014open}. MIMU has been proven to be useful in improving accuracy for positioning, bias, and coarse alignment \cite{larey2020multiple}, and has various applications such as calibration \cite{rehder2016extending, carlsson2021self}, integration with GNSS \cite{guerrier2009improving}, pedestrian navigation \cite{bancroft2010multiple, skog2014pedestrian, bose2017noise}, data fusion and filtering \cite{patel2022multi, patel2021sensor, bancroft2011data}, and localization algorithms \cite{zhang2020lightweight}.

In recent years, machine and deep learning (DL) approaches have revolutionized the inertial sensing field.
Cohen \emph{et al}. provided a comprehensive review of various approaches to applying deep learning to inertial sensing and sensor fusion \cite{cohen2024inertial}. Direct bias estimation using a DL approach was initially addressed in \cite{engelsman2022learning} and indirect gyroscope calibration in \cite{huang2022mems,gao2022gyro}. The ability of DL to process complex data and learn intricate patterns makes it a powerful tool for improving navigation systems and enhancing their accuracy and reliability in real-world scenarios.

This study introduces a neural network-based approach to improve the zero-order calibration of low-cost gyroscopes by leveraging real and virtual data. Our deep learning method significantly reduced calibration time while improving accuracy compared to the baseline model-based approach. Specifically, using a single IMU, our approach demonstrated notable improvements in both accuracy and calibration time. Incorporating virtual data further enhanced these gains, while training with multiple gyroscopes provided additional benefits, particularly in reducing calibration time. These results remained consistent across different gyroscope brands, reinforcing the robustness of our approach.
In this work, we integrate our approach with multi-gyroscope (MG) data to significantly reduce gyroscope calibration time while maintaining high accuracy. Unlike previous methods that rely solely on a single gyroscope or extensive calibration durations, our approach leverages multiple real and virtual gyroscopes to enhance bias estimation.
Beyond its quantitative improvements, this method has significant practical implications. In applications where rapid calibration is crucial, such as robotics, autonomous vehicles, and search and rescue operations, our approach provides a viable alternative to traditional calibration methods, which require longer stationary periods. The ability to integrate virtual data also reduces the need for large-scale real-world sensor datasets, making this approach more scalable and adaptable to various hardware configurations.
By bridging the gap between conventional calibration techniques and data-driven methods, our study contributes to the advancement of rapid and efficient inertial sensor calibration, with broad potential applications in navigation, robotics, and wearable technology.
This approach is particularly beneficial in time-sensitive applications, such as search and rescue operations \cite{davids2002urban}, aerial vehicles \cite{bernard2011autonomous}, or air quality measurements using drones {\cite{bakirci2024smart}}. Although it may not directly improve precision, it enables quicker deployment and adaptation of the navigation systems, allowing these applications to integrate and use new gyroscopes more rapidly.
To evaluate our proposed approach, we recorded 76.2 hours from 36 gyroscopes of four different brands. In addition, we created a virtual dataset of 110.4 hours. Our real recorded dataset is publicly available at the GitHub repository\footnote{https://github.com/ansfl/RapidGyroCalibration} , to allow reproducibility of our work and to increase research in the field.
Our results show improvements over the model-based baseline approach both in accuracy and in rapid calibration.

The rest of this paper is organized as follows: Section \ref{sec:problem_formulation} describes the model-based calibration approach; Section \ref{proposed_approach} presents our proposed approaches; Section \ref{results} provides an analysis and discussion, and Section \ref{conclusions} concludes this paper.
\section{Formulation of the Problem} \label{sec:problem_formulation}
In our study, we focused on calibrating low-cost gyroscopes in stationary conditions. The gyroscope error model is \cite{Groves2013}:
 \begin{align} \label{eq:gyro_error_model}
\boldsymbol{\tilde{\omega}}^g = \mathbf{M} \boldsymbol{\omega}^g + b_g + \boldsymbol{w_g} 
\end{align}
where $\boldsymbol{\tilde{\omega}}_{g}$ is the gyroscope measurement expressed in the gyro frame g, $\boldsymbol{\omega}_{g}$ is the true angular velocity vector expressed in frame g, $\mathbf{M}$ is a matrix of the misalignment (off-diagonal elements) and scale factor (diagonal terms) errors, $\boldsymbol{b}_g$ is the gyroscope bias, and $\boldsymbol{w}_g$ is zero mean white Gaussian noise.
The most common technique for sensor calibration is zero-order calibration, which involves taking measurements while the gyroscope is stationary. In the zero-order calibration method, the bias is estimated by taking the mean over a sequence of stationary measurements.
The underlying assumption is that the measurement noise is zero-mean and the actual measurement should be zero, as low-cost gyroscopes are not capable of measuring the earth rotation rate (low signal-to-noise ratio). Thus, taking the expectation operator from both sides of \eqref{eq:gyro_error_model} gives:
 \begin{equation} \label{eq:average_on_gyro}
\mathbb{E}[\boldsymbol{\tilde{\omega}}^b_{ib}] = \textbf{M} \cancelto{0}{\boldsymbol{\omega}^b_{ib}} + b_g + \cancelto{\sim0}{\mathbb{E}[\boldsymbol{w_g}]} \hspace{0.8cm} \triangleq b_{g,d}
\end{equation}
where $b_{g,d}$ is the deterministic part of the gyroscope bias.
Equation \eqref{eq:average_on_gyro} demonstrates how averaging the gyroscope measurements in stationary conditions results in an estimate of the bias only. The performance of this calibration approach depends on the number of measurements and therefore on the calibration time.
\begin{figure}[!htbp]
\begin{center}
\captionsetup{justification=centering}
\includegraphics[scale=0.5]{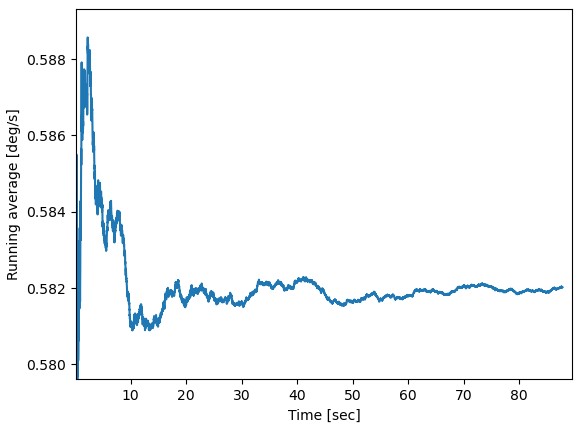}
\caption{Example of a running average  applied to measurements over time of a stationary gyroscope until convergence is achieved, approximately after 63 seconds.}
\label{fig:running_average}
\end{center}
\end{figure}
Figure \ref{fig:running_average} shows an example of a running average on a stationary gyroscope readings over time until convergence is achieved, approximately after 63 seconds. 
The value after the convergence is addressed as the gyroscope bias. 

During operation (after the calibration stage is completed), the estimated deterministic bias is subtracted from the gyroscope measurement:
\begin{equation} \label{eq:gyro_cal}
    \boldsymbol{\tilde{\omega}}_{c}^g = \boldsymbol{\omega}^g - b_{g,d}
\end{equation}
where $\boldsymbol{\tilde{\omega}}_{c}^g$ represents the calibrated gyro measurements.
\begin{figure}[!htbp]
\begin{center}
\captionsetup{justification=centering}
\includegraphics[scale=0.5]{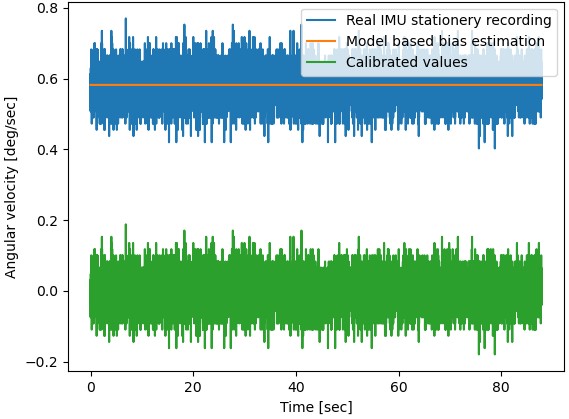}
\caption{Raw and calibrated gyroscope signals. The blue signal presents the raw stationery gyroscope measurements, the orange line being the deterministic bias. The calibrated gyro measurements are shown in the green signal.}
\label{fig:CalibratedVsUncalibrated}
\end{center}
\end{figure}

We continue our example in Figure \ref{fig:running_average} and present the calibrated gyroscope readings in Figure \ref{fig:CalibratedVsUncalibrated}. The blue signal shows the raw stationery gyroscope measurements with the deterministic bias being the orange line. The calibrated gyro measurements are shown in the green signal. As expected, the calibration lowered the gyroscope readings to their expected value.
\section{Proposed Approach} \label{proposed_approach}
Our goal is to improve the performance of gyroscope calibration using neural networks. To further enhance our approach, we used real MGs and a virtual MG array to assist in the calibration of a single IMU consisting of three gyroscopes. 
We proposed two DL calibration methods: one with increasing input channels and the other with increasing the training data with real and virtual MGs.

In simple terms, our goal is to enhance the accuracy of gyroscope signals by leveraging artificial intelligence (AI). Specifically, we use deep learning, a branch in AI, to predict the inherent errors in the gyroscope measurements and compensate for them, thereby improving overall sensor performance. By applying deep learning techniques, our approach can model complex error patterns that standard calibration methods may not fully capture. This enables precise and reliable motion estimation, which is particularly beneficial for applications such as navigation, robotics, and autonomous systems.
\subsection{Motivation}\label{subsec:MG_calibration}
\begin{figure}[!htbp]
\begin{center}
\captionsetup{justification=centering}
\includegraphics[scale=0.1]{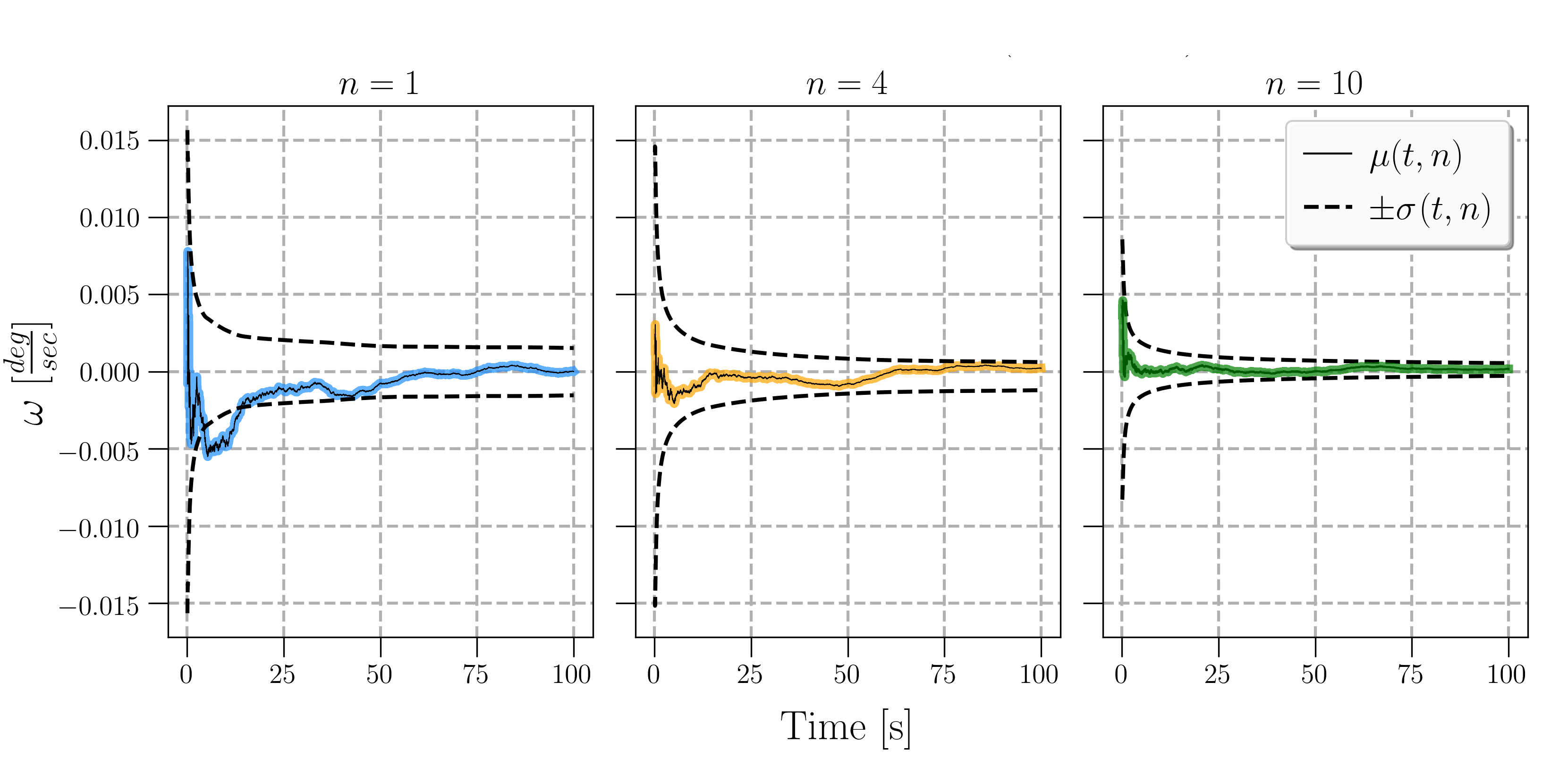}
\caption{MG convergence over time as a function of the number of gyroscopes. (Left): Using a single gyroscope. (Center): Using 4 gyroscopes. (Right): Using 10 gyroscopes.}
\label{fig:mg_calibration}
\end{center}
\end{figure}

When considering multiple IMUs operating together in close proximity, commonly a virtual IMU is used \cite{libero2024augmented} to average all the physical inertial readings from existing IMUs into a single virtual one. In this case, MG calibration leverages multiple gyroscopes to achieve superior error estimation compared to that of single gyroscope. By integrating data from several gyroscopes, each from the same series, the MG setup operates as a unified system, thereby refining the precision of the sensor data. Figure \ref{fig:mg_calibration} illustrates the running average convergence behavior of MG stationary bias calibration over time, showing the effect of increasing the number of gyroscopes on the accuracy of the running average of the calibrated values.  The results demonstrate that as the number of gyroscopes incorporated into the MG setup increases, the mean value converges more rapidly and closely to the ground truth, reflecting an improvement in calibration speed and precision. 
Motivated by the use of MG to improve calibration performance, in this research, we integrated DL algorithms with MG to achieve rapid and accurate gyroscope calibration.
\subsection{Increasing Input Channels}\label{subsec:raising_in_ch}
The basic NN approach uses single gyroscope readings as input to output its bias. We increased the number of input channels,  each channel representing a single gyroscope so that the output size corresponds to the number of channels (gyroscopes). As an IMU consists of three orthogonal gyroscopes, with the addition of each IMU, the input channels are increased by three. We sought to train the network with multiple channels and rely on the interconnections between the MG readings to improve the calibration performance. 
For example, when training the network with data from three gyroscopes, the network input consists of three channels corresponding to the three axes and
the network output is the three deterministic bias values. Next, we examined the effect of increasing the number of input channels by incorporating data from additional gyroscopes, resulting in an input of 3N channels, where N is the number of IMUs. Note that the training data increases with the use of addition gyroscopes. If M is the number of samples recorded by each gyro, the training data with consist of $3 \cdot N \cdot M$ samples.  

\begin{figure}[!htbp]
\begin{center}
\captionsetup{justification=centering}
\includegraphics[scale=0.3]{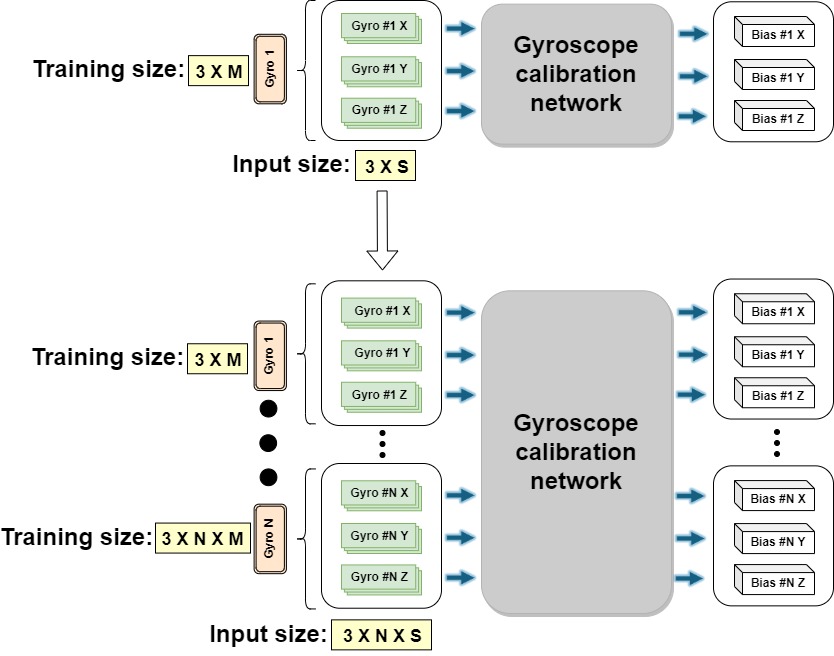}
\caption{Block diagram of our proposed approach to increasing the number of input channels. The upper diagram depicts a setup with three gyroscopes while the lower has $3\dot N$ gyroscopes. The figure shows the differences in the training data size, input channels, and output. N is the number of IMUs, M is the number of samples recorded by each gyroscope, and S is the window size.}
\label{fig:raising_input_ch}
\end{center}
\end{figure}

Figure \ref{fig:raising_input_ch} provides a graphic illustration of our proposed approach and demonstrates how incorporating additional gyroscopes increases the training data volume, input channels, and output.

\subsection{Increasing Training Data}\label{subsec:raising_train_data}
\begin{figure}[!htbp]
\begin{center}
\captionsetup{justification=centering}
\includegraphics[scale=0.3]{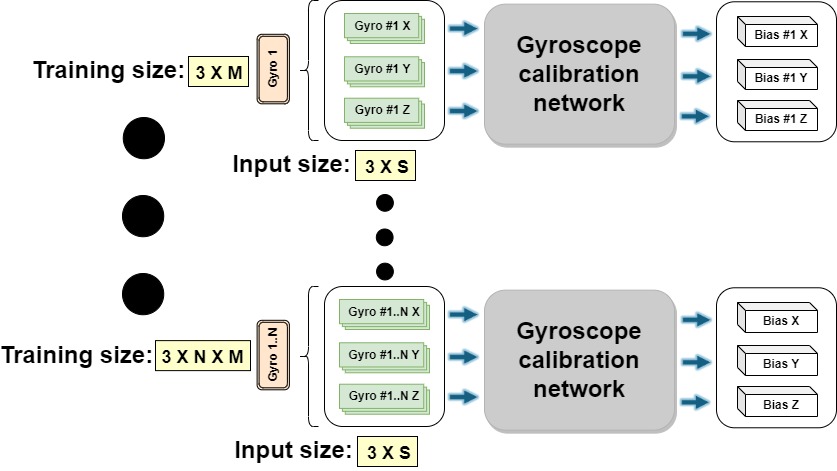}
\caption{Block diagram illustrating our approach for increasing the training data. The upper diagram depicts a setup with three gyroscopes; the lower diagram shows $3\dot N$ gyroscopes. The figure shows the differences in the training data size. N is the number of IMUs, M is the number of samples recorded by each gyroscope, and S is the window size.}
\label{fig:raising_train_data}
\end{center}
\end{figure}

Recall that the basic NN approach uses single gyroscope readings as input to output its bias. Following this approach, we used a fixed number of input channels and increased the training data using real and virtual MGs. Given that an IMU consists of three orthogonal gyroscopes, we set the input channels to three. We sought to implement a basic DL principle according to which when increasing the training data the performance of NNs improves until reaching a steady state solution, that is, until additional data no longer influences performance. Therefore, increasing the training data should improve the calibration performance of the gyroscope, up to a steady-state solution (saturation). For example, when training the network with data from three gyroscopes, the minimum training set consists of $3 \cdot M$ samples.  When increasing the training set by an additional $N$ IMUs, the training set consists of $3 \cdot M \cdot N$ samples.  The input channels, however, remain three in those two examples. Figure \ref{fig:raising_train_data}  provides a graphic illustration of our approach and demonstrates how incorporating additional gyroscopes increases the volume of the training data while maintaining the size of the input channels.

\subsection{Neural Network Architecture}
\begin{figure}[!htbp]
\begin{center}
\captionsetup{justification=centering}
\includegraphics[scale=0.3]{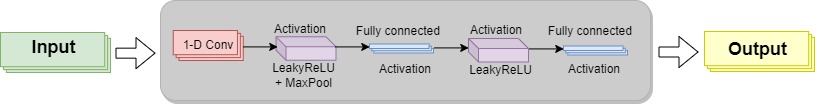}
\caption{Our baseline network architecture, suitable for both proposed approaches. The network receives gyroscope measurements and outputs the gyroscope deterministic bias.}
\label{fig:learning_approach}
\end{center}
\end{figure}

Aiming to use the same NN in both proposed approaches, we conducted initial experiments with various neural architectures. 
These included convolutional layers and recurrent neural networks such as long short-term memory layers.  Based on this evaluation, we decided to focus on convolutional neural network architecture, as presented in Figure \ref{fig:learning_approach}. \\
It consists of a convolution layer followed by a LeakyReLU activation function \cite{maas2013rectifier} and a max-pooling layer. Next, two fully connected layers, with a LeakyReLU activation function between them, process the output features. Convolutional layers were chosen due to their ability to efficiently extract spatial and temporal patterns from gyroscope data while maintaining computational efficiency. Unlike RNNs or transformers, CNNs require fewer parameters and are less prone to overfitting when applied to relatively short, structured time series. Additionally, LeakyReLU was selected over standard ReLU to prevent issues related to dying neurons, ensuring stable gradient flow during training.
The input to the network is a multi-channel time series data from multiple gyroscopes. The input to the first layer is:
\begin{equation}
    \mathbf{X} \in \mathbb{R}^{3 \cdot N \times S}
\end{equation}
where $N$ is the number of gyroscopes and $S$ is the window size. In the $i^{th}$ convolutional layer, the output $Z_{i}$ is obtained as follows:
\begin{equation} 
    Z_i =  \sum_{j=1}^{m} (x_j + (j-1)s \cdot w_j) + b
\end{equation}
where $m$ represents the window/kernel size, $b$ is the bias, $s$ denotes the stride, and $w_j$ represent the weights. The $\sigma_{\text{LeakyReLU}}$ is the LeakyReLU activation function, defined as follows:
\begin{equation}
  \sigma_{\text{LeakyReLU}}(Z_i) = \begin{cases} 
        Z_i, & \text{if } Z_i \geq 0 \\
        \alpha \cdot Z_i, & \text{otherwise}
        \end{cases}
\end{equation}
where \( \alpha \) (slope) is 0.1. Then, we apply max pooling to $Z_i$:
\begin{equation}
    Y_i = \max_{m=1}^{P} Z_{i \cdot s + m}
\end{equation}
where $P$ is the pooling size. After flattening the input into a 2-dimensional tensor, we computed the first fully connected layer:
\begin{equation}
    \boldsymbol{L_1} = Y_i \boldsymbol{W_1} + \boldsymbol{b_1}
\end{equation}
where $\boldsymbol{W_1}$ and $\boldsymbol{b_1}$ are the weights and biases of the first fully connected layer. We applied the activation function and repeated the process for the second fully connected layer, resulting in the final output $\hat{y}_i$:
\begin{equation}
    \hat{y}_i = \sigma_{\text{LeakyReLU}}(\boldsymbol{L_1}) \boldsymbol{W_2} + \boldsymbol{b_2}
\end{equation}
where $\boldsymbol{W_2}$ and $\boldsymbol{b_2}$ are the weights and biases of the second fully connected layer, respectively.

For the training process, we used the mean squared error (MSE) loss function:
\begin{equation}
\text{MSE} = \frac{1}{n} \sum_{i=1}^{n} (y_i - \hat{y}_i)^2
\end{equation}
where $n$ is the number of data points, $y_i$ is the actual bias value, and $\hat{y}_i$ is the predicted bias value. 
The GT bias is obtained using a long-duration recording, whereas all approaches were evaluated over shorter periods, ranging from 5\% to 50\% of the total time required for the GT. We used the Adam optimizer \cite{kingma2014adam} to train the neural network, leveraging its adaptive learning rate and efficient gradient-based optimization to enhance convergence speed and accuracy. The batch size was 64, with a learning rate of 0.0001 and a learning rate decay of 0.1 every 200 epochs. Training was conducted over 1,200 epochs.
\section{Analysis and Results} \label{results}
We begin by describing the gyroscopes used in our experiments and the corresponding datasets. Next, we outline our evaluation process for both proposed approaches and compare them with the model-based baseline. Following this, we present the results and conclude with a brief summary of our findings.
\subsection{Dataset} \label{subsec:dataset}
We employed four types of gyroscopes for our experiments: (a) Movella Dot \cite{MovellaDot}, (b) SparkFun \cite{SparkFun}, (c) NG-IMU {\cite{NG}} and (d) Memsense MS-IMU3025 {\cite{Memsense}}. The specifications for these gyroscopes, as provided by the manufacturers, are shown  in Table \ref{sensor_specifications}.

\begin{table}[htbp]
    \centering
    \caption{Specifications of the gyroscopes, as provided by the manufacturers.}
    \begin{adjustbox}{max width=0.5\textwidth}
        \begin{tabular}{|l|c|c|c|}
            \hline
            \textbf{Sensor name} & \textbf{Sample rate} [Hz] & \textbf{Noise density} [mdps/$\sqrt{\text{Hz}}$] & \textbf{Bias stability}    [$\deg$/h] \\
            \hline
            Movella DOT & 120 & 7 & 10 \\
            \hline
            SparkFun & 130-145 & 3.8 & N/A \\
            \hline
            NG & 200 & N/A & N/A \\
            \hline
            Memsense & 250 & 14.8 & 2.6 \\
            \hline
        \end{tabular}
    \end{adjustbox}
    \label{sensor_specifications}
\end{table}

We used four Movella DOTs and four SparkFun IMUs in our experiments. After evaluating our approach on those devices, we further tested our approach on the NG and Memsense IMUs to evaluate its generalization. During data acquisition, all 12 IMUs from each brand were placed on a stable table to minimize external disturbances.
For synchronization, different methods were applied based on sensor type. Movella DOT sensors were synchronized using built-in Bluetooth functionality, allowing simultaneous activation and recording via a mobile application. For SparkFun, NG, and Memsense IMUs, which do not have built-in synchronization, we designed and 3D-printed custom placeholders to keep the sensors in a fixed position and connected them to a single power hub. A centralized data acquisition script, controlled by Arduino, was implemented to trigger data logging simultaneously across all connected sensors, ensuring time alignment of the recorded gyroscope readings.
To account for potential variations in gyroscope performance due to manufacturing inconsistencies, we conducted experiments using multiple units of each IMU type. Despite being from the same manufacturer, individual sensors exhibited slight differences in bias and noise characteristics. By including multiple sensors in our dataset and evaluating performance across different models, we ensured that our approach remains robust to these variations.

To achieve an accurate bias estimate from a single recording, a substantial number of samples is necessary to average out the Gaussian mean-zero noise. Consequently, 13,000 samples were recorded for each of the gyroscopes. The bias distribution was estimated by repeating the experiment 100 times. The training dataset should include a variety of bias values to allow generalization. Consequently, we turned off the device and waited 10 seconds before turning it back on. By doing so, we obtained different bias values in each experiment, which contributed to the training process and allowed for generalization.
In addition to the real-world dataset, we generated virtual gyroscope measurements using simulation. The goal of this dataset was twofold: (a) increase the training set data for better results and (b) examine whether less data recorded by real gyroscopes, combined with virtual data, may be used to achieve similar performance in the bias regression task. This process was done only for the DOT and SparkFan.
To create the virtual dataset, we first calculated the bias ground truth values of the 24 gyroscopes (both brands) using \eqref{eq:average_on_gyro}. To ensure that the virtual gyroscope simulation accurately represents real gyroscope characteristics, we derived bias values from the measured uniform distribution of real gyroscopes. The results of this analysis provided an estimate for a reasonable standard deviation range. Each IMU type was assigned a standard deviation value based on the noise characteristics observed in its real recordings, with $0.04$ [$\frac{deg}{sec}$] for the SparkFun gyroscopes. Finally, we generated gyroscope measurements from this normal distribution. To this end,  given a bias value, we generated a sequence of 13,000 virtual measurements for a single recording. In total, for the same bias value, we made 100 recordings. We randomly selected  24 bias values, each representing a virtual gyroscope, and generated 100 recordings for each of them.
\begin{figure}[!htbp]
\begin{center}
\captionsetup{justification=centering}
\includegraphics[scale=0.2]{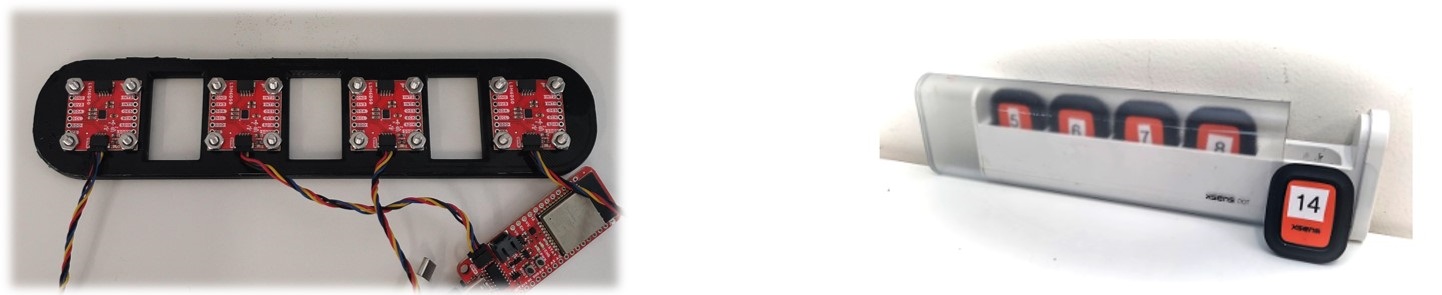}
\caption{Experimental setup: (right) Movella DOTS IMUs configuration, (left) SparkFun IMUs configuration.}
\label{fig:MG_config}
\end{center}
\end{figure}

For the evaluation process, we divided the real and virtual data into four datasets:
\begin{itemize}
    \item \textbf{Dataset-1}: Contains recorded data from all 12 SparkFun gyroscopes across 100 recordings, each with 13,000 measurements corresponding to 87 seconds of recording time per session. The training set includes 23.2 hours of recording and the testing set includes 1.45 hours of recording that are not present in the training.
    
    \item \textbf{Dataset-2}: Includes data from all 12 DOT gyroscopes across 400 recordings, each with 13,000 measurements corresponding to 120 seconds of recording time per session. The training set includes 32 hours of recording and the testing set includes 2 hours of recording that are not present in the training.

    \item \textbf{Dataset-3}: Contains virtual data from 24 virtual SparkFun gyroscopes with a total time of 46.4 hours. This dataset is used only as additional training data for Dataset-1.
    
    \item \textbf{Dataset-4}: Includes virtual data from 24 virtual DOT gyroscopes with a total time of 64 hours. This dataset is used only as additional training data for Dataset-2.

    \item \textbf{Dataset-5}: The dataset contains 100 recordings of each of the 6 NG gyroscopes. Recordings consist of 13,000 measurements lasting 65 seconds. Training set recordings include 8.67 hours, while testing set recordings include 1.08 hours not included in training set recordings.

    \item \textbf{Dataset-6}: The dataset contains 100 recordings of each of the 6 Memsense gyroscopes. Recordings consist of 13,000 measurements lasting 52 seconds. Training set recordings include 6.83 hours, while testing set recordings include 0.87 hours not included in training set recordings.
\end{itemize}
Table \ref{tab:dataset_in_hours} shows the duration of all six datasets. Overall, the {six} datasets contain 186.6 hours of gyroscope measurements from 36 real gyroscopes and 48 virtual ones. \\
\begin{table}[!htbp] 
    \centering 
    \caption{Detailed description of the total training and testing time across all datasets.}
    \begin{adjustbox}{max width=1\textwidth}
        \begin{tabular}{|c|c|c|c|}
            \hline
            \textbf{Dataset} & \textbf{Train [hours]} & \textbf{Test [hours]} & \multicolumn{1}{l|}{\textbf{Total [hours]}} \\ \hline
            1 & 23.2 & 1.45 & 24.65 \\ \hline
            2 & 32 & 2   & 34   \\ \hline
            3 & 46.4 & 0 & 46.4 \\ \hline
            4 & 64 & 0   & 64   \\ \hline
            5 & 8.67 & 1.08   & 9.75   \\ \hline
            6 & 6.93 & 0.87   & 7.8   \\ \hline
        \end{tabular}
    \end{adjustbox}
    \label{tab:dataset_in_hours}
\end{table}
\subsection{Evaluation Metric and Approach} \label{subsec:evaluation}
We chose the root mean square error (RMSE) metric to quantify the performance of the model-based method and of our learning approaches. The RMSE is defined as follows:

\begin{equation}
    \text{RMSE} = \sqrt{\frac{1}{N} \sum_{i=1}^{N} (y_i - \hat{y}_i)^2}
    \label{eq:rmse}
\end{equation}

where $N$ is the number of predicted biases, $y_i$ represents the GT bias from the test dataset, and $\hat{y}_i$ is the estimated bias. 

From a machine learning perspective, more data in the training set improves performance. Practically, it is much easier to create a virtual dataset than to record data from MGs. Therefore, we used three training approaches based on the same test dataset:

\begin{itemize}
    \item \textbf{Real2Real}: Only real recorded data were used for training, either from Dataset-1, Dataset-2, Dataset-5 or Dataset-6. In this approach we examined a minimum of three gyroscopes (a single IMU) and a maximum of 12 gyroscopes (4 IMUs).

    \item \textbf{(3 Real+Virtual)2Real}: This approach involved mixing Dataset-1 with Dataset-3 and Dataset-2 with Dataset-4. Only the data from three gyroscopes (a single IMU) was used in the real dataset, with additional data from the virtual sets.

    \item \textbf{(12 Real+Virtual)2Real}: This approach involved mixing Dataset-1 with Dataset-3 and Dataset-2 with Dataset-4. The data from 12  gyroscopes was used in the real dataset, with additional data from the virtual sets.
    
\end{itemize}
The testing dataset for all three approaches contained real recorded data from Dataset-1 or Dataset-2. We repeated this procedure for testing the other IMUs. We found that a different set of parameters was required (with the same network architecture) for each IMU. Thus, without loss of generality and for the sake of simplicity of the presentation, we focus only on the three gyroscopes of IMU $\#1$.
\subsection{Increasing the Number of Input Channels} \label{subsec:increase_ch_method}
\begin{figure}[!htbp]
\begin{center}
\captionsetup{justification=centering}
\includegraphics[scale=0.5]{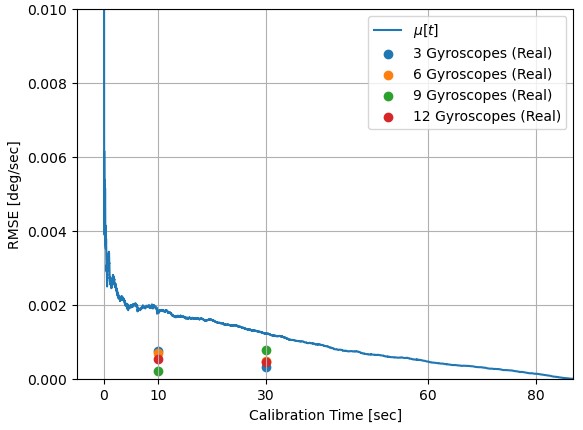}
\caption{Method 1 demonstrates the effects of increasing the number of input channels during each neural network training session using the SparkFun gyroscopes. All the results were obtained on the same testing dataset. The results indicate an improvement in the performance of the neural network compared to the model-based approach but with inconsistent behavior.}
\label{fig:raise_in_ch_results}
\end{center}
\end{figure}

In this section, we explore the effect of increasing the number of input channels and training data during neural network training on gyroscope calibration accuracy, using the Real2Real approach. Specifically, we analyze the performance of the neural network when using data from Dataset-1. 
We computed the running RMSE across the test dataset using the model-based approach. Subsequently, we trained the network four times with varying numbers of gyroscopes in the input to the network and training set. We repeated this process twice, once with a 10-second calibration period and again with a 30-second period. We chose calibration times that represent approximately 10\% and 33\% of the time required by the model-based approach to converge. From a data-driven point of view, we wanted to give the network significant time to learn and therefore used 10 seconds as a minimum. To meet the inertial requirements, we needed rapid calibration and, thus, limited the maximum time to 30 seconds. Finally, we determined the time difference between the RMSE achieved by the neural network after 10 or 30 seconds and the time required for the model-based approach to reach the same RMSE.\\
Figure \ref{fig:raise_in_ch_results} illustrates the results of the increasing input channels method. The blue line represents the running RMSE average obtained from the model-based approach using the test dataset. By contrast, the colored dots depict the RMSE of the neural network for varying numbers of gyroscopes. The graph shows an improvement in calibration time, although the overall performance still lags behind that of the model-based approach. We anticipated that adding more gyroscopes would lead to an improvement in RMSE but despite adjusting the parameters, the results did not show stable improvement. 
\subsection{Increasing the Size of the Training Dataset} \label{subsec:increase_train_method}
\begin{figure}[!htbp]
\begin{center}
\captionsetup{justification=centering}
\includegraphics[scale=0.5]{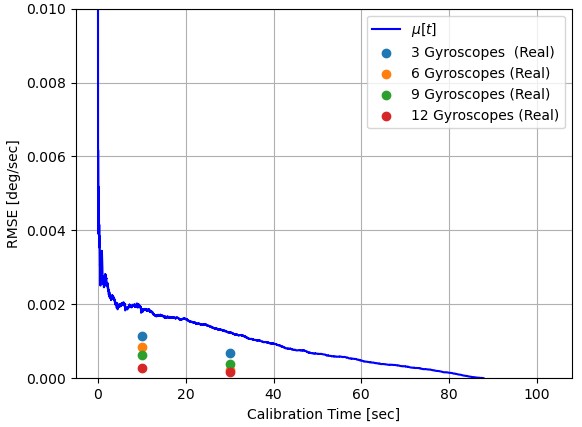}
\caption{Method 2: Increasing the SparkFun training data. The results demonstrate the effect of expanding the training dataset. Our approach shows rapid calibration and improved bias accuracy. All the results were obtained on the same testing dataset.}
\label{fig:raise_train_data_results}
\end{center}
\end{figure}

\begin{figure}[!htbp]
\begin{center}
\captionsetup{justification=centering}
\includegraphics[scale=0.4]{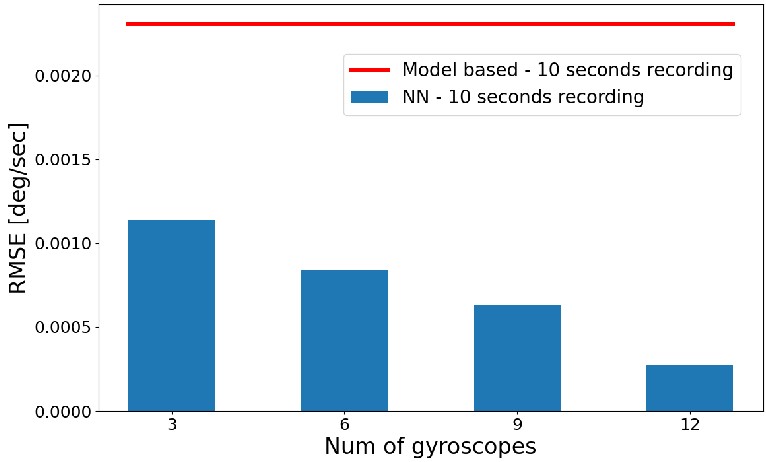}
\caption{Results for the ten second calibration time as a function of the number of gyroscopes in the training set using SparkFun gyroscopes. All the results were obtained on the same testing dataset. This closer look provides insights into the efficiency and accuracy of the calibration over a short time frame, indicating the capability of the method to quickly correct biases.}
\label{fig:raise_train_data_10_sec}
\end{center}
\end{figure}

\begin{figure}[!htbp]
\begin{center}
\captionsetup{justification=centering}
\includegraphics[scale=0.5]{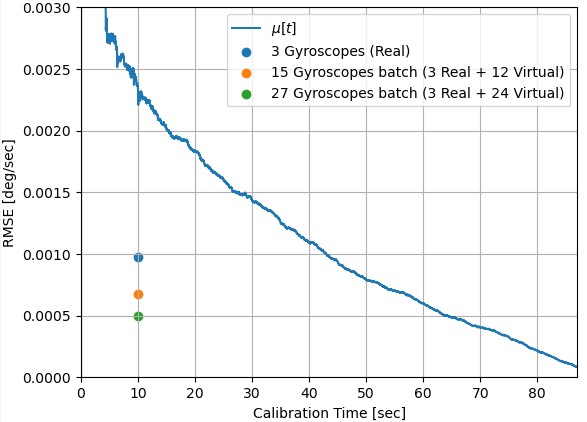}
\caption{Augmenting SparkFun training data by combining it with virtual data from Dataset-3. The results demonstrate the effect of expanding the training dataset. Our approach shows rapid calibration and improved bias accuracy. All the results were obtained on the same testing dataset.}
\label{fig:real_plus_sim_raise_train_data}
\end{center}
\end{figure}

We followed the same procedure as in the previous section and examined the influence of the size of the training dataset on the calibration performance. We began the evaluation with three gyroscopes in the training set, and each time, we increased the number of gyroscope recordings by three until we reached 12 gyroscopes. Figure \ref{fig:raise_train_data_results} presents the results of our analysis. The blue line in the figure represents the running RMSE average obtained from the model-based approach using the test dataset (the same as in Figure \ref{fig:raise_in_ch_results}).  By contrast, the colored dots depict the RMSE of the neural network for varying numbers of gyroscopes. Figures {\ref{fig:raise_train_data_results}}, {\ref{fig:real_plus_sim_raise_train_data}}, {\ref{fig:merged_real_sim_raise_train_data}}, {\ref{fig:dots_real_plus_sim_raise_train_data}}, and {\ref{fig:merged_dots_real_plus_sim_raise_train_data}} follow the same concept as Figures {\ref{fig:raise_in_ch_results}} and {\ref{fig:raise_train_data_results}}, illustrating similar results but using different datasets or methods, as described in Sections {\ref{subsec:evaluation}}, {\ref{subsec:increase_ch_method}}, and {\ref{subsec:increase_train_method}}. We note that as additional gyroscopes are added to the train dataset, the performance improves both for the 10- and the 30-second calibration time. Regardless of the number of gyroscopes, we obtain rapid convergence relative to the model-based approach. For example, when using three gyroscopes, it takes the model-based approach 39 seconds to reach the performance that it takes 10 seconds to achieve in our approach.  
Figure \ref{fig:raise_train_data_10_sec} provides a closer look at the results achieved with the 10-second calibration time. The blue bars represent the RMSE achieved with our approach using varying numbers of gyroscopes, while the red line corresponds to the model-based approach. This figure provides a closer view of the ten-second calibration results presented in Figure \ref{fig:raise_train_data_results}. In this case, not only a raid calibration was achieved, but also, for the same calibration time, our calibration network improved upon the accuracy of the model-based approach. For example, when using three gyroscopes, the improvement was 57\% and when using 12 gyroscopes, the improvement increased to 88\%.
\subsection{Adding Virtual Data} \label{sec:add_sim_data}
We first examined the possibility of improving a set of three gyroscopes (single IMU) using a virtual dataset. To this end, we applied the increasing the size of the training set approach and added virtual gyroscope readings using the (Real+Virtual)2Real method. We first trained on the data of three real gyroscopes from Dataset-1 and added virtual gyroscope readings from Dataset-3 to the training set. We repeated this process until the input included 27 gyroscopes, only three being real. Regardless of the size of the training set, the testing dataset was from Dataset-1.
The results indicate that using 27 virtual gyroscopes led to 84\% improvement in calibration time, decreasing the model-based calibration time from 64 to 10 seconds. This approach allows us to rely on virtual data instead of collecting a large dataset from MGs.

\begin{figure}[!htbp]
\begin{center}
\captionsetup{justification=centering}
\includegraphics[scale=0.5]{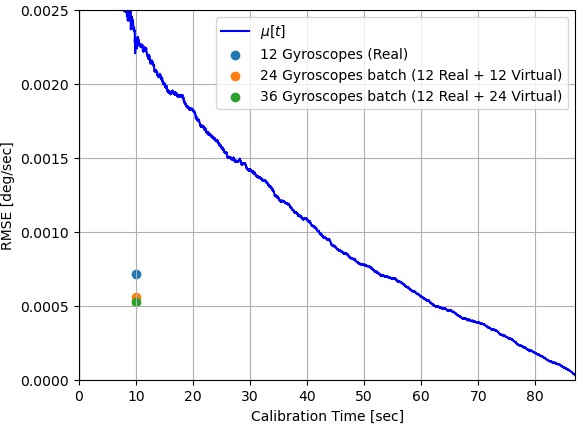}
\caption{Augmenting SparkFun training data by merging real and virtual data from Dataset-3. The results demonstrate the effect of expanding the training dataset. Our approach shows rapid calibration and improved bias accuracy. All the results were obtained on the same testing dataset.}
\label{fig:merged_real_sim_raise_train_data}
\end{center}
\end{figure}

We conducted a further evaluation to assess the influence of MG by merging all 12 real gyroscopes with virtual data following the (12 Real+Virtual)2Real approach. In the first training session, the dataset consisted of 12 real gyroscopes from Dataset-1. In the subsequent session, 12 virtual gyroscopes from Dataset-3 were added, resulting in a total of 24 gyroscopes (12 real + 12 virtual). This process was continued until the dataset included 36 gyroscopes: 24 virtual and 12 real. The results, shown in Figure \ref{fig:merged_real_sim_raise_train_data}, demonstrate even better outcomes. For example, using 24 virtual gyroscopes led to 84\% improvement in calibration time, by decreasing the model-based calibration time from 62 seconds to 10. A steady improvement was observed as more virtual data were added in both methods.

\subsection{Approach Robustness} \label{subsec:approach_robustness}
\begin{figure}[!htbp]
\begin{center}
\captionsetup{justification=centering}
\includegraphics[scale=0.5]{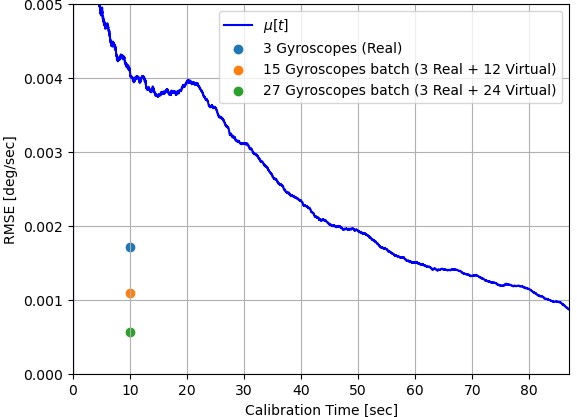}
\caption{Augmenting Movella DOTs training data by combining real and virtual data from Dataset-4. 
The results demonstrate the effect of expanding the training dataset. Our approach shows rapid calibration and improved bias accuracy. All the results were obtained on the same testing dataset.}
\label{fig:dots_real_plus_sim_raise_train_data}
\end{center}
\end{figure}

To examine the robustness of our approach with different types of IMUs, we used the DOT IMU recordings. We followed the same procedure as in the previous section, and examined the influence of the training dataset on calibration performance, using virtual data. We used the same algorithm as before but this time with DOT gyroscopes.
The real data were taken from Dataset-2 and the virtual data from Dataset-4. We began the evaluation with three gyroscopes (single IMU) in the training set and increased the number of gyroscope recordings by 12 virtual gyroscopes until we reached 27 gyroscopes (3 Real + 24 Virtual). Figure \ref{fig:dots_real_plus_sim_raise_train_data} presents the results of our analysis. 
The results show that regardless of gyroscope type, we achieved rapid convergence compared to the model-based approach. For example, when using three gyroscopes it took the model-based approach 54 seconds to reach the performance that we achieved with our approach in 10 seconds. When adding 24 virtual gyroscopes it raises to  an 89\% improvement in  calibration time. 

Next, we examined performance with MG using 12 DOT gyroscopes. We evaluated the model by initially using 12 real gyroscopes in the training set, gradually increasing the number of gyroscopes by incorporating 12 additional virtual gyroscopes at each step, until reaching a total of 36 gyroscopes (12 Real + 24 Virtual).
\begin{figure}[!htbp]
\begin{center}
\captionsetup{justification=centering}
\includegraphics[scale=0.5]{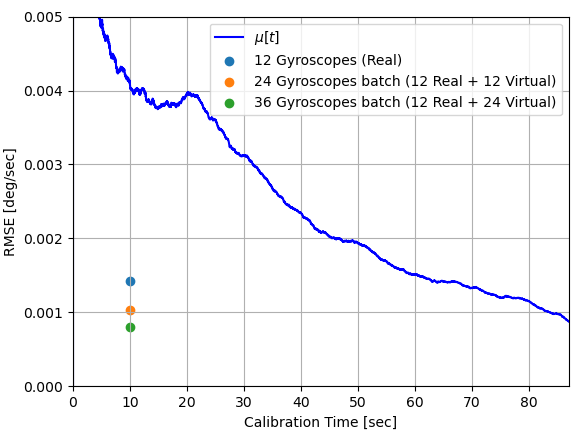}
\caption{Augmenting Movella DOTs training data by merging real and virtual gyroscope data from Dataset-4. 
The results demonstrate the effect of expanding the training dataset. Our approach shows rapid calibration and improved bias accuracy. All results were obtained using the same testing dataset.}
\label{fig:merged_dots_real_plus_sim_raise_train_data}
\end{center}
\end{figure}
Figure \ref{fig:merged_dots_real_plus_sim_raise_train_data} illustrates the outcomes of this analysis. The results attest to the advantages of using MG across different types of gyroscopes. For example, when using 12 DOTs gyroscopes, the model-based approach required 67 seconds to achieve the performance that we obtained using  our approach in 10 seconds. With the addition of 24 virtual gyroscopes,  an 88\% improvement in calibration time was achieved.

\subsection{Summary} \label{subsec:summary}
In {\cite{engelsman2022learning}} a DL approach was used to estimate single gyroscope zero order calibration. While presenting novelty in using DL for gyro calibration, this approach couldn’t achieve better performance for the same calibration time. 
As demonstrated, using our Real2Real and (Real+ Virtual)2Real approaches, the results consistently indicate significant improvements in calibration time and accuracy. Table \ref{tab:three_gyro_results} summarizes the results using three gyroscopes (single IMU) and virtual data. In all the approaches, we improved the accuracy of the model-based approach for short calibration times. Additionally, the model-based approach requires more time to achieve the performance that we achieve with a 10-second calibration interval. We demonstrated this performance using two types of IMUs. The addition of virtual data improved both accuracy and calibration time compared to using only three real gyroscopes.
\begin{table}[!htbp]
\centering
\captionsetup{justification=centering}
\caption{Summary of calibration results using three real gyroscopes with and without virtual data in a 10-second calibration time.}
\begin{adjustbox}{max width=0.5\textwidth}
\begin{tabular}{|c|c|cc|}
\hline
\multirow{2}{*}{Method}     & \multirow{2}{*}{Number of gyroscopes} & \multicolumn{2}{c|}{\begin{tabular}[c]{@{}c@{}}Three gyroscopes (single IMU)\\ Improvement using our approach\end{tabular}}                                                                                            \\ \cline{3-4} 
                            &                                    & \multicolumn{1}{c|}{\begin{tabular}[c]{@{}c@{}}Calibration time \\ (same performance) {[}\%{]}\end{tabular}} & \begin{tabular}[c]{@{}c@{}}Accuracy improvement \\ (same calibration time) {[}\%{]}\end{tabular} \\ \hline
Real2Real                   & 3 Real                             & \multicolumn{1}{c|}{72}                                                                                             & 57                                                                                               \\
(3 Real + Virtual)2Real     & 3 Real + 24 Virtual                & \multicolumn{1}{c|}{84}                                                                                             & 79                                                                                               \\
(3 DOT Real + Virtual)2Real & 3 Real + 24 Virtual                & \multicolumn{1}{c|}{89}                                                                                             & 85                                                                                               \\ \hline
\end{tabular}
\end{adjustbox}
\label{tab:three_gyro_results}
\end{table}

Table \ref{tab:mg_results} shows our extended analysis incorporating MG data, presenting the calibration results as in the previous table. To this end, we used the readings of 12 real gyroscopes (4 IMUs) in the training process. The results demonstrate that both calibration time and performance showed marked improvements when MG data were used. Training on real data alone contributed significantly to the reduction in convergence time, and  the integration of virtual data further enhanced both time efficiency and overall performance metrics. Yet, training on real data alone, without incorporating virtual data, produced greater accuracy improvements for the same calibration time.
\begin{table}[!htbp]
\centering
\captionsetup{justification=centering}
\caption{Summary of calibration results using MG (12 real gyroscopes) with and without virtual data in a 10-second calibration time.}
\begin{adjustbox}{max width=0.5\textwidth}
\begin{tabular}{|c|c|cc|}
\hline
\multirow{2}{*}{Method}       & \multirow{2}{*}{Num of gyroscopes} & \multicolumn{2}{c|}{\begin{tabular}[c]{@{}c@{}}Three gyroscopes (single IMU)\\ improvement using our approach\end{tabular}}                                                                                            \\ \cline{3-4} 
                              &                                    & \multicolumn{1}{c|}{\begin{tabular}[c]{@{}c@{}}Calibration time \\ (same performance) {[}\%{]}\end{tabular}} & \begin{tabular}[c]{@{}c@{}}Accuracy improvement \\ (same calibration time) {[}\%{]}\end{tabular} \\ \hline
Real2Real                     & 12 Real                            & \multicolumn{1}{c|}{86}                                                                                             & 88                                                                                               \\
(12 Real + Virtual)2Real      & 12 Real + 24 Virtual               & \multicolumn{1}{c|}{84}                                                                                             & 77                                                                                               \\
(12 DOTs Real + Virtual)2Real & 12 Real + 24 Virtual               & \multicolumn{1}{c|}{88}                                                                                             & 80                                                                                               \\ \hline
\end{tabular}
\end{adjustbox}
\label{tab:mg_results}
\end{table}
\begin{table}[!htbp]
\centering
\captionsetup{justification=centering}
\caption{Summary of calibration results using three real gyroscopes without virtual data in a 10-second calibration time.}
\begin{adjustbox}{max width=0.5\textwidth}
\begin{tabular}{|c|c|cc|}
\hline
\multirow{2}{*}{IMU type}     & \multirow{2}{*}{Number of gyroscopes} & \multicolumn{2}{c|}{\begin{tabular}[c]{@{}c@{}}Three gyroscopes (single IMU)\\ Improvement using our approach\end{tabular}}                                                                                            \\ \cline{3-4} 
                            &                                    & \multicolumn{1}{c|}{\begin{tabular}[c]{@{}c@{}}Calibration time \\ (same performance) {[}\%{]}\end{tabular}} & \begin{tabular}[c]{@{}c@{}}Accuracy improvement \\ (same calibration time) {[}\%{]}\end{tabular} \\ \hline
SparkFan                   & 9 Real                             & \multicolumn{1}{c|}{82}                                                                                             & 87                                                                                               \\
DOT     & 9 Real                & \multicolumn{1}{c|}{89}                                                                                             & 82                                                                                               \\
NG & 6 Real                & \multicolumn{1}{c|}{83}                                                                                             & 98                                                                                               \\ 
Memsense & 6 Real                & \multicolumn{1}{c|}{75}                                                                                             & 75          \\ \hline
\end{tabular}
\end{adjustbox}
\label{tab:generalization}
\end{table}
Table {\ref{tab:generalization}} presents the results for the four IMU types using the Real2Real Method with increasing training data. The results demonstrate our approach's versatility across different IMU types. Three of the IMU manufacturers examined with our approach achieved calibration time and accuracy results of 80\% or higher. Using the Memsense IMU the improvement was 75\%. These results highlight our method's effectiveness.
\section{Conclusions} \label{conclusions} 
This study introduced a neural network-based approach to improve the zero-order calibration of low-cost gyroscopes by leveraging real and virtual data. Our deep learning method significantly reduced calibration time while improving accuracy compared to the baseline model-based approach. Specifically, using a single IMU, our approach achieved a 57\% improvement in accuracy and a 72\% reduction in calibration time. Incorporating virtual data further enhanced accuracy by 84\% and reduced calibration time by 79\%. When training with 12 MGs, the method showed substantial improvements in both metrics, with virtual data primarily contributing to faster calibration. These results remained consistent across different gyroscope brands, reinforcing the robustness of our approach.

Beyond its quantitative improvements, this method has significant practical implications. In applications where rapid calibration is crucial, such as robotics, autonomous vehicles, and search and rescue operations, our approach provides a viable alternative to traditional calibration methods, which require longer stationary periods. The ability to integrate virtual data also reduces the need for large scale real world sensor datasets, making this approach more scalable and adaptable to various hardware configurations. By bridging the gap between conventional calibration techniques and data-driven methods, our study contributes to the advancement of rapid and efficient inertial sensor calibration, with broad potential applications in navigation, robotics, and wearable technology.

However, our findings reveal a critical trade-off between calibration speed and accuracy. With no time constraints, the model-based baseline approach continues to provide the most precise bias estimation. Yet, in applications such as search and rescue and robotics, where rapid calibration is a must, our approach has critical mission advantages. Therefore, users can choose the most suitable method based on their concrete requirements, balancing the need for speed against accuracy and enhancing both flexibility and efficiency in various practical scenarios. Future work could explore extending this approach to dynamic calibration scenarios, improving real-time implementation, and optimizing the neural network for embedded systems to further enhance its applicability in resource-constrained environments.

\section*{Acknowledgement}
Y. S. receives support from the Maurice Hatter scholarship.
\bibliographystyle{IEEEtran}
\bibliography{Ref}
\end{document}